\pdfoutput=1

\documentclass[11pt]{article}

\usepackage{emnlp2021}

\usepackage{times}
\usepackage{latexsym}

\usepackage[T1]{fontenc}

\usepackage[utf8]{inputenc}
\usepackage{graphicx}

\usepackage{microtype}

\usepackage[normalem]{ulem}
\usepackage{enumitem}
\usepackage{bbding}
\makeatletter
\def\blfootnote{\gdef\@thefnmark{}\@footnotetext}
\makeatother
\title{Project Debater APIs: Decomposing the AI Grand Challenge} 

\author{ Roy Bar-Haim, Yoav Kantor, Elad Venezian$^{*}$, Yoav Katz, Noam Slonim\\
IBM Research\\
\texttt{\{roybar,yoavka,eladv,katz,noams\}@il.ibm.com}}

\begin{document}
\maketitle
\begin{abstract}
\emph{Project Debater} was revealed in 2019 as the first AI system that can debate human experts on complex topics. Engaging in a live debate requires a diverse set of skills, and Project Debater has been developed accordingly as a collection of components, each designed to perform a specific subtask. \emph{Project Debater APIs} provide access to many of these capabilities, as well as to more recently developed ones. This diverse set of web services, publicly available for academic use, includes core NLP services, argument mining and analysis capabilities, and higher-level services for content summarization. We describe these APIs and their performance, and demonstrate how they can be used for building practical solutions. In particular, we will focus on \emph{Key Point Analysis}, a novel technology that identifies the main points and their prevalence in a collection of texts such as survey responses and user reviews.

\end{abstract}
\section{Introduction}
\blfootnote{
    \hspace{-0.2cm}
		$^{*}$First three authors equally contributed to this work.
}
Argumentation and debating are fundamental capabilities of human intelligence. They are essential for a wide range of  everyday activities that involve reasoning, decision making or persuasion. Over the last few years, there has been growing interest in \emph{Computational Argumentation}, defined as ``the application of computational methods for analyzing and synthesizing argumentation and human debate'' \citep{gurevych_et_al:DR:2016:5803}.  A recent milestone in this field is \emph{Project Debater}, which was revealed in 2019 as the first AI system that can debate human experts on complex topics\footnote{\url{https://www.research.ibm.com/artificial-intelligence/project-debater/}}.
Project Debater is the third in the series of IBM Research AI’s grand challenges, following Deep Blue and Watson. It has been developed for over six years by a large team of researchers and engineers, and its live demonstration in February 2019 received massive media attention. In our recent paper, “An autonomous debating system”, published in the Nature magazine \cite{Slonim2021}, we describe Project Debater’s architecture and evaluate its performance.

To debate humans, an AI must be equipped with a diverse set of skills. It has to be able to pinpoint relevant arguments for a given debate topic in a massive corpus, detect the stance of arguments and assess their quality. It also has to identify principled, recurring arguments that are relevant for the specific topic, organize the different types of arguments into a compelling narrative, recognize the arguments made by the human opponent, and make a rebuttal. Accordingly, Project Debater has been developed as a collection of components, each designed to perform a specific subtask. Over the years, we published more than 50 papers describing these components and released many related datasets for academic use.

Successfully engaging in a debate requires high level of accuracy from each component. For example, failing to detect the argument’s stance may result in arguing in favor of your opponent – a dire situation in a debate. A crucial part of developing highly accurate models was the collection of uniquely large scale, high-quality labeled datasets for training each component. The evidence detection classifier, for instance, was trained using 200K labeled examples, and was able to achieve a remarkable precision of 95\% for top 40 candidates \citep{EinDor-2020}.

Another major challenge was scalability. One example is applying Wikification \citep{mihalcea:2007} to our 10 billion sentences corpus, a task that was infeasible for any of the available tools. We therefore developed a novel, fast Wikification algorithm, which can be applied to massive corpora while achieving competitive accuracy \cite{shnayderman2019fast}. 

\emph{Project Debater APIs} give access to selected capabilities originally developed for the live debating system, as well as related technologies we have developed more recently.  We provide free access for academic use to these APIs, as well as trial and licensing options for developers. The APIs can be divided into three main groups:
\begin{itemize}[topsep=4pt,itemsep=1pt,parsep=4pt]
\itemsep0.1em 
\item \emph{Core NLU services}, including Wikification, semantic relatedness between Wikipedia concepts, short text clustering, and common theme extraction for texts. These general-purpose tools may be useful in many different use cases, and may serve as building blocks in a variety of NLP applications.
\item \emph{Argument Mining and Analysis}, including the detection of sentences containing claims and evidence, claim boundaries detection within a sentence, argument quality assessment and stance classification (pro/con). These services are of particular interest to the computational argumentation research community. 
\item \emph{Content summarization}, including two high-level services: \emph{Narrative Generation} constructs a well-structured speech that supports or contests a given topic, according to the specified polarity. \emph{Key Point Analysis} summarizes a collection of comments  as a small set of automatically extracted, human-readable key points, each assigned with a numeric measure of its prominence in the input. These tools may serve data scientists analyzing opinionated texts such as user reviews, survey responses, social media, customer feedback, etc.    
\end{itemize}

Several demonstrations of argument mining capabilities have been previously published \citep{stab-etal-2018-argumentext, wachsmuth-etal-2017-building, chernodub-etal-2019-targer}, some of which also provide access to their capabilities via APIs. However, Project Debater APIs offer a much broader set of services, trained on unique large-scale, high quality datasets, which have been developed over many years of research.

The next sections describe each of the APIs and their performance assessment, and how they can be accessed and used via the Debater Early Access Program. We then describe several examples of using and combining these APIs in practical applications.
\section{Services Overview}
In this section we provide a short description for each service, and point to its related publications, and other relevant resources. All the training datasets for these services have been developed as part of Project Debater.
\subsection{Core NLU Services}
This group of services includes several fundamental natural language processing tasks.

\paragraph*{Text wikification.}
The Wikification service identifies mentions of Wikipedia concepts in the given text. We created our own wikifier, described in \cite{shnayderman2019fast}, since existing tools were far too slow to be applied to the Lexis-Nexis corpus we used for argument mining, which contains about 10 billion sentences. We developed a simple rule-based method, which relies on matching the mentions to the Wikipedia title, as well as on Wikipedia redirects.  This approach enables very fast Wikification, about 20 times faster than the commonly-used TagMe Wikifier \citep{Ferragina:2010}, while achieving competitive accuracy.

\paragraph*{Concept relatedness.}
This service  measures the semantic relatedness between a pair of Wikipedia concepts. We trained a BERT regressor \cite{bert-2019} on the WORD dataset \cite{EinDor:2018}, which includes  
13K pairs of Wikipedia concepts manually annotated to determine their level of relatedness. The input to the regressor is the first sentence in the Wikipedia article of each concept. 

\paragraph*{Text clustering.}
Our Text clustering service is based the Sequential Information Bottleneck (sIB)  algorithm \cite{Slonim-SIB}. This unsupervised algorithm has been shown to achieve strong results on standard benchmarks. However, sIB has not been as popular as other clustering algorithms such as K-Means, since its run time was significantly higher.  Our implementation of sIB is highly optimized, leveraging the sparseness of bag of words representation. With this optimization, the run time of sIB is very fast, and even comparable with K-Means. The python code of this implementation is also available\footnote{\url{https://github.com/IBM/sib}}.

\paragraph*{Common theme extraction.}
This service gets a clustering partition of sentences and returns Wikipedia concepts representing the main themes in each cluster. These themes aim to represent the subjects that are discussed by the sentences of this cluster, and distinguish it from other clusters. This service is based on the hypergeometric test, applied to the concepts mentioned in the sentences of each cluster. The service identifies concepts that are enriched in each cluster compared to the other clusters, taking into account the semantic relatedness of different concepts.
\subsection{Argument Mining and Analysis}
This group includes classifiers and regressors that aim to identify arguments in input texts, determine their stance, and assess their quality. 

\paragraph*{Claim Detection.}
This service identifies whether a sentence contains a claim with respect to a given topic. This task was introduced by \citet{levy-etal-2014-context}. They define a Claim as \emph{``a general, concise statement that directly supports or contests
the given Topic''}. The claim detection model is a BERT-based classifier, trained on 
90K positive and negative labeled examples from the Lexis-Nexis corpus. The model is similar to  the one described in \cite{EinDor-2020}.
\paragraph*{Claim Boundaries.}
Given an input sentence that is assumed to contain a claim, this service returns the boundaries of the claim within the sentence \citep{levy-etal-2014-context}.
The Claim Boundaries service may be used to refine the results of the Claim Detection service, which provides sentence-level classification. The service is based on a BERT model, which was fine-tuned on  
52K crowd-annotated examples mined from the Lexis-Nexis corpus.
\paragraph*{Evidence Detection.}
Similar to the Claim Detection service, this service gets a sentence and a topic and identifies whether the sentence is an Evidence supporting or contesting the topic. In our context, an Evidence is an argument that contains research results or an expert opinion. This is a BERT based service which was fine-tuned using 
200K annotated examples from Lexis-Nexis corpus. This model is based on the work of \citet{EinDor-2020}.
\paragraph*{Argument Quality.}
This service, based on the work of  \citet{Gretz_Friedman_Cohen-Karlik_Toledo_Lahav_Aharonov_Slonim_2020}, produces a numeric quality score for a given argument. The service is based on a BERT regressor, which was trained on 27K arguments, collected for a variety of topics and annotated with quality scores. Both the arguments and the quality scores were collected via crowdsourcing. The real-valued argument quality scores were derived from a large number of binary labels collected from crowd annotators. Specifically, for each example, the annotators were asked whether the sentence, as is, may fit in a speech supporting or contesting the given topic. High quality scores typically indicate arguments that are grammatically valid, use proper language, make a clear and concise argument, have a clear stance towards the topic, etc.

\paragraph*{Pro-Con.}
This service \citep{Barhaim:2017, toledo-ronen-etal-2020-multilingual}, gets an argument and a topic and predicts whether the argument supports or contests the topic. This service is a BERT-based classifier, which was trained on 400K stance-labeled examples. It includes arguments extracted from the Lexis-Nexis corpus, as well as arguments collected via crowsourcing. The set of training arguments was automatically expanded by replacing the original debate concept with consistent and contrastive expansions, based on the work of \citet{bar-haim-etal-2019-surrogacy}. 

\subsection{Content Summarization}
This group contains two high-level services that create different types of summaries.

\paragraph*{Key Point Analysis.}
 This service summarizes a collection of comments on a given topic as a small set of \emph{key points} \cite{bar-haim-etal-2020-arguments, bar-haim-etal-2020-quantitative}. The salience of each key point is given by the number of its matching sentences in the given comments.  The input for the service is a collection of textual comments, which are split into sentences. The output is a short list of key points and their salience, along with a list of matching sentences per key point. A key point matches a sentence if it captures the gist of the sentence, or is directly supported by a point made in the sentence. The service selects key points from a subset of concise, high-quality sentences (according to the quality service described above), aiming to achieve high coverage of the given comments. Matching sentences to key points is performed by a RoBERTa-large model \cite{roberta-2019}, trained on a dataset of 24K (argument, key point) pairs, labeled as matched/unmatched. It is also possible to specify the key points as part of the input, in which case the service matches the sentences to the given key points. 

\paragraph*{Narrative Generation}
This service receives a topic, and a list of arguments that support or contest the topic, and constructs a well-structured speech summarizing the relevant input arguments that are compatible with the requested stance (pro or con). 

It works as follows: first, we select high-quality arguments with the right stance. Then, the service performs Key Point Analysis over these arguments. Finally, The service selects the most prominent key points, and for each key point, it selects the best arguments to create a corresponding paragraph. Alternatively, paragraphs may be generated based on the output of the text clustering service. Selected arguments are slightly rephrased as required and connecting text is added to improve the fluency of the resulting speech. 

\subsection{Wikipedia Sentence-Level Index}
In addition to the above groups of services, we also provide a sentence-level index of Wikipedia.
The index underlying our search service is a data structure that is populated with sentences, enriched with some metadata, such as the Wikipedia concepts mentioned in each sentence (as identified by the Wikification service), named entities, and multiple lexicons. The index facilitates fast retrieval of sentences according to queries that may refer to the text and/or the metadata, with word distance restrictions.   For example, retrieve all the sentences that satisfy the template  ``\texttt{<PERSON> … that … <CONCEPT> …  <SENTIMENT-WORD>}''.

\section{Assessment}
Table \ref{tab:assessment} includes assessment results for various services. For each service, we specify the benchmark that was used for testing, the evaluation measure(s) and the results. If the results in the table are quoted from one of our papers, this is indicated by \Checkmark. Unless otherwise mentioned, the results are from the same paper that is cited for the dataset. In cases where the results for the service were not available (this happens, for example, if the current service implementation is different from the one described in the paper), we ran the service on the benchmark and reported the results.

The text clustering assessment is the only one that is not performed over a Project Debater dataset, but over a standard benchmark - the widely-used 20 newsgroups dataset, which contains about 18,000 news posts on 20 topics \cite{LANG1995331}. We clustered these posts into 20 clusters, and compared the results with the original partition. We report Adjusted Mutual Information (AMI) and Adjusted Rand Index (ARI) measures. Our results (AMI=0.595 and ARI=0.466) are considerably better than those obtained with K-Means (AMI=0.228 and ARI=0.071).

Overall, the results confirm the high quality of our services.
\begin{table*}[h!]
\begin{small}
\begin{center}
\begin{tabular}{ |p{0.17\textwidth}|p{0.3\textwidth}|p{0.1\textwidth}|p{0.1\textwidth}|c| } 
 \hline
 \textbf{Service} & \textbf{Benchmark} & \textbf{Measure} & \textbf{Result} & \textbf{From Paper?} \\
 \hline
 \hline
 Evidence Detection & VLD \cite{EinDor-2020}& Precision@40 & 0.95 &  \Checkmark\\
 \hline
 Argument Quality & IBM-Rank30k-WA \cite{Gretz_Friedman_Cohen-Karlik_Toledo_Lahav_Aharonov_Slonim_2020} & Pearson / Spearman correlation & 0.52 / 0.48 &  \Checkmark\\
 \hline
 Concepts  relatedness & WORD  \cite{EinDor:2018}& Pearson / Spearman correlations & 0.85 / 0.57 &\\
 \hline
 Text  wikification & Trans  \cite{shnayderman2019fast}& Precision / Recall / F1 & 0.76 / 0.62 / 0.68 &  \Checkmark\\
 \hline
 Pro-Con & IBM-Rank30k \cite{Gretz_Friedman_Cohen-Karlik_Toledo_Lahav_Aharonov_Slonim_2020}  &Accuracy & 0.92 & \\
 \hline
 Text  clustering \ & 20 Newsgroups \citep{LANG1995331} & AMI / ARI  &  0.595 / 0.466 & \\
 \hline
 Key Point Analysis & ArgKP \citep{bar-haim-etal-2020-arguments}; Results are from \citep{bar-haim-etal-2020-quantitative}& F1 & 0.77 &  \Checkmark\\
 \hline
\end{tabular}
\caption{Project Debater APIs assessment\label{tab:assessment}}
\end{center}
\end{small}
\end{table*}

\section{The Debater Early Access Program}
The 12 Project Debater APIs are offered via the \emph{IBM Debater Early Access Program}. The goal of this program is to make core capabilities from Project Debater available as building blocks for a variety of text understanding applications. 

The Early Access Program is freely available for academic use on the IBM Cloud, and can also be licensed for commercial use. As part of the program, both Python and Java SDKs are available. All the services are REST-based, which enables their usage by any desired programming language.
In order to access these APIs, an API key is required. We supply such API keys freely for non-commercial use.

The early access website, shown in Figure~\ref{fig:screen_shot}, contains various resources\footnote{\url{https://early-access-program.debater.res.ibm.com/academic_use}}. The main tab includes a detailed description of all the services with Python, Java, and CURL code examples. The description also includes links to related publications. In addition, it contains a demo UI, which allows interacting with the APIs online.

The \emph{Examples} tab contains step-by-step tutorials, which demonstrate how the APIs can be applied in complex scenarios, to solve real-world problems. The \emph{Data Sets} tab contains a link to all Project Debater datasets. Finally, there is a tab for the \emph{Speech By Crowd} application. Speech By Crowd is a web application that enables the user to collect and analyze opinions on a desired controversial topic. The application is free for non-commercial use.  The application has been demonstrated in several events, as we discuss in the next section.

\begin{figure*}
    \centering
    \includegraphics[width=\textwidth]{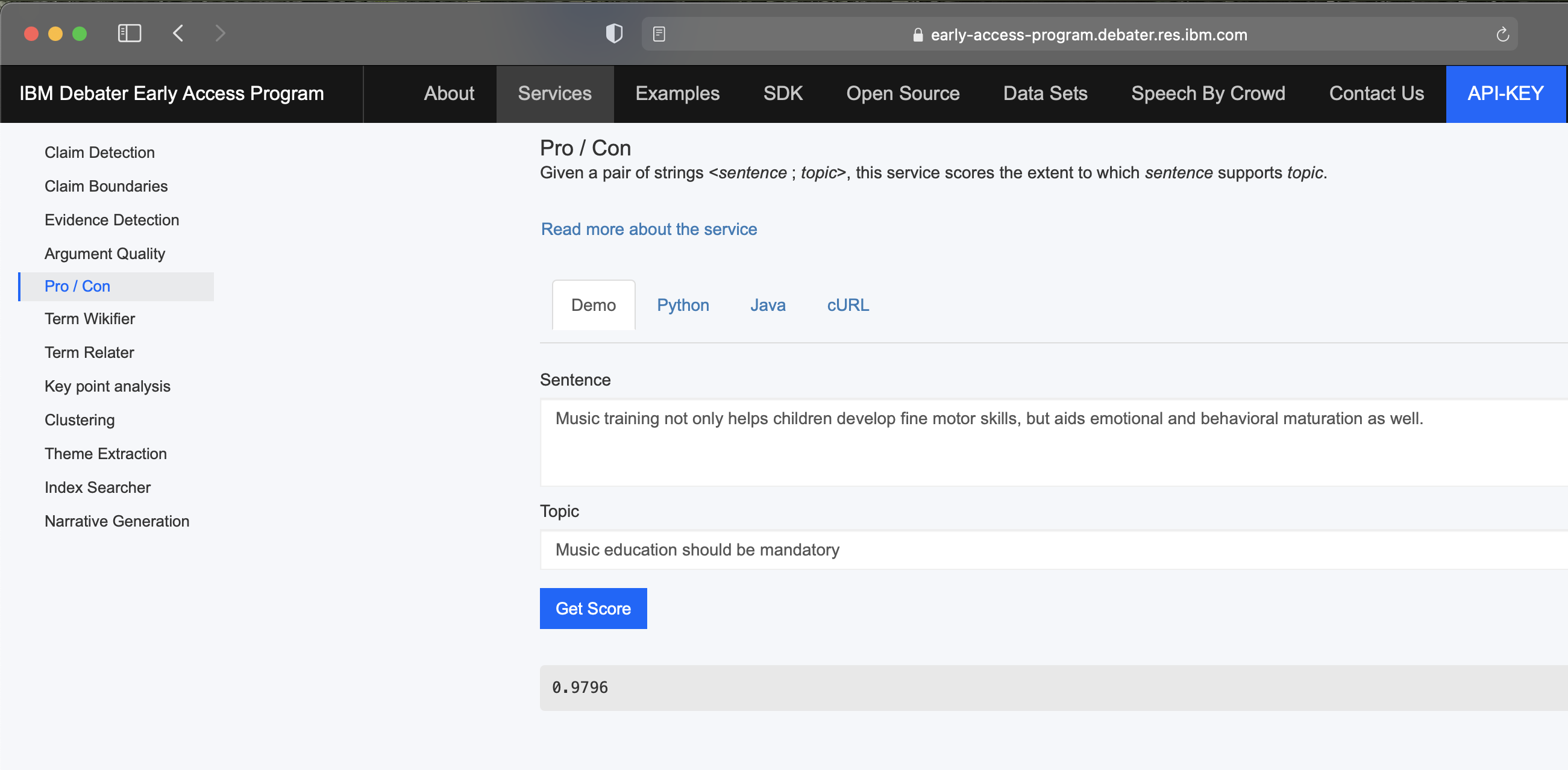}
    \caption{The IBM Debater Early Access Program web page. The list of services is shown on the left. For the selected service - the pro-con service, there is an expandable short description and a demo that allows trying the service online.}
    \label{fig:screen_shot}
\end{figure*}

\section{Use Cases}
\subsection{Analysing Surveys and Reviews}
\begin{figure*}[t]
    \centering
    \includegraphics[width=\textwidth]{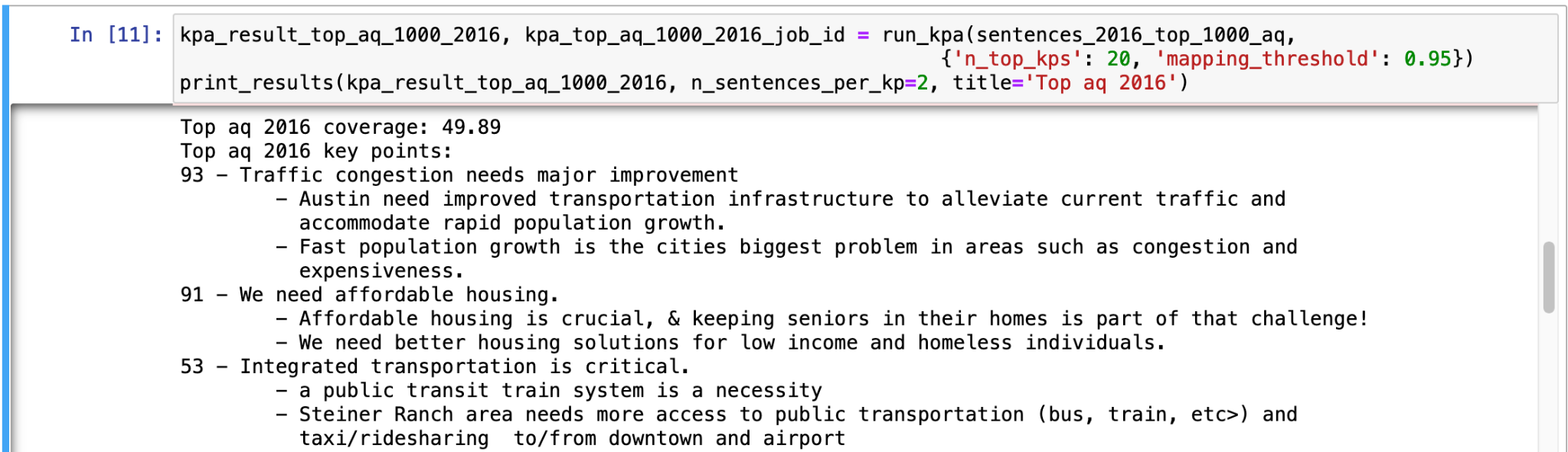}
    \caption{A screenshot from the Jupyter Notebook of the tutorial, showing the results for running KPA on the 2016 Austin survey data (1000 top-quality sentences). The coverage, top three key points and top 2 matching sentences per key point are displayed.}
    \label{fig:screen_shot}
\end{figure*}

Surveys are commonly used by decision makers to collect opinions from a large audience. However, extracting the key issues that came up in hundreds or thousands of survey responses is a very challenging task. 

Existing automated approaches are often limited to identifying key phrases or concepts and the overall sentiment toward them, but do not provide detailed, actionable insights. Using Debater APIs, and in particular Key Point Analysis (KPA), we are able to analyze and derive insights from answers to open-ended survey questions. 

\paragraph{Austin Municipal Survey Tutorial.} To demonstrate this capability, we have prepared a hands-on tutorial, publicly available on GitHub\footnote{\url{https://github.com/IBM/debater-eap-tutorial}}. In this tutorial, we analyze free-text responses for a community survey conducted in the city of Austin in the years 2016 and 2017. 
In this survey, the citizens of Austin were asked ``If there was ONE thing you could share with the Mayor regarding the City of Austin (any comment, suggestion, etc.), what would it be?''. 

In the tutorial, we first run KPA on 1000 randomly selected sentences from 2016. 
We then use the Argument Quality (AQ) service and run KPA on 1000 top-quality sentences from 2016. We show that selecting higher-quality sentences for our sample results in a better summary, and higher coverage of the resulting key points. Figure~\ref{fig:screen_shot} is a screenshot from the Jupyter Notebook of the tutorial. It shows the overall coverage of the extracted key points, and lists the top key points found, and for each key point - the number of its matching sentences and its top-scoring matches. 

The tutorial also shows how to compare the 2016 responses to those  from 2017. This can be done by mapping 1000 top-quality sentences from 2017 to the same set of key points that was extracted for 2016, and observe the year-to-year changes in the key point salience. 

The results show that traffic congestion is one of the top problems in Austin. In order to better understand the citizens' complaints and suggestions regarding this topic, we can use two additional services, Wikification and Concept Relatedness, to identify sentences that are related to the \emph{Traffic} concept and run KPA only on this subset. 

\paragraph*{IBM Employee Engagement Survey.}
Key Point Analysis has also been applied to analyze the 2020 IBM employee engagement survey. Over 300K employees wrote more than 550K sentences in total. These sentences were automatically classified into positive and negative, and we ran KPA on each set separately to extract positive and negative key points.  
The HR team reported that these analyses enable them to extract actionable and valuable insights with significantly less effort.

\paragraph*{Business Reviews.}
Similar to surveys, KPA can also be used for effectively summarizing user reviews. In our recent work \citep{barhaim2021bite} we demonstrate its application to the Yelp dataset of business reviews. 

\subsection{Online Debates}
In the following public demonstrations, we combined several services to summarize online debates, where hundreds or thousands of participants submit online their pro and con arguments for a controversial topic, using the Speech by Crowd platform. We used the \emph{pro-con} service to split arguments by stance, the \emph{argument quality} service to filter out low quality arguments, the \emph{KPA} service to summarize the data into key points and the \emph{narrative generation} to create a coherent speech.
\paragraph*{That's Debateable.}
``That’s Debatable'' is a TV show presented by Bloomberg Media and Intelligence Squared. In each episode,  a panel of experts debates a controversial topic, such as \emph{``It’s time to redistribute the wealth''}. Using the above pipeline, we were able to summarize thousands of arguments submitted online by the audience, and the resulting pro and con key points and speeches were presented during the show. The audience contributed interesting points, some of which were not raised by the expert debaters, and therefore enriched the discussion.\footnote{\url{https://www.research.ibm.com/interactive/project-debater/thats-debatable/}}

\paragraph*{Grammy Music Debates.} During the Grammys 2021 event, four music debate topics (e.g., virtual concerts vs. live shows) were published on the event's website. Hundreds of arguments contributed by music fans were collected for each topic, and the same method was applied to analyze and summarize them\footnote{\url{https://www.grammy.com/watson}}.

\section{Conclusion}
We introduced \emph{Project Debater APIs}, which provide access to many of the core capabilities of the Project Debater grand challenge, as well as more recent technologies such as Key Point Analysis. The evaluation we presented confirms the high quality of these services. We discussed different use cases for these APIs, in particular for analyzing and summarizing various types of opinionated texts. We believe that this diverse set of services may be used as building blocks in many text understanding applications, and may be relevant for a broad audience in the NLP community. 
\section*{Acknowledgements}
The authors thank Alon Halfon, Naftali Liberman, Amir Menczel,  Guy Moshkowich, Dafna Sheinwald, Ilya Shnayderman and  Artem Spector for their contribution to the development of the IBM Debater Early Access Program. 
\bibliography{custom}
\bibliographystyle{acl_natbib}
\end{document}